\documentclass[10pt,journal,compsoc]{IEEEtran}
\ifCLASSOPTIONcompsoc
  \usepackage[nocompress]{cite}
\else
  \usepackage{cite}
\fi
\ifCLASSINFOpdf
\else
\fi

\usepackage{times}
\usepackage{epsfig}
\usepackage{graphicx}
\usepackage{amsmath}
\usepackage{amssymb}
\usepackage[ruled,vlined]{algorithm2e}
\usepackage{algpseudocode} 
\usepackage{multirow}

\usepackage{url}
\usepackage[pagebackref=true,breaklinks=true,colorlinks,bookmarks=false]{hyperref}

\begin{document}
%
\title{PLSM: A Parallelized Liquid State Machine for Unintentional Action Detection}

\author{Dipayan Das, 
        Saumik Bhattacharya, 
        Umapada Pal,~\IEEEmembership{Senior Member,~IEEE,}
        and Sukalpa Chanda. 

\IEEEcompsocitemizethanks{\IEEEcompsocthanksitem D. Das is with Computer Vision and Pattern Recognition Unit, \\Indian Statistical Institute, Kolkata, India.\protect\\
E-mail: dipayan.das2010@gmail.com

\IEEEcompsocthanksitem S. Bhattacharya is with Dept. of Electronics and Electrical Communication Engg, IIT Kharagpur, India\protect\\
E-mail: saumik@ece.iitkgp.ac.in.
\IEEEcompsocthanksitem U. Pal is with Computer Vision and Pattern Recognition Unit, \\ Indian Statistical Institute, Kolkata, India.\protect\\
E-mail: umapada@isical.ac.in
\IEEEcompsocthanksitem S. Chanda is with Østfold University College, Norway.\protect\\
E-mail: sukalpa@ieee.org}}
\IEEEtitleabstractindextext{%
\begin{abstract}
Reservoir Computing (RC) offers a viable option to deploy AI algorithms on low-end embedded system platforms. Liquid State Machine (LSM) is a bio-inspired RC model that mimics the cortical microcircuits and uses spiking neural networks (SNN) that can be directly realized on neuromorphic hardware. In this paper, we present a novel Parallelized LSM (PLSM) architecture that incorporates spatio-temporal read-out layer and semantic constraints on model output. To the best of our knowledge, such a formulation has been done for the first time in literature, and it offers a computationally lighter alternative to traditional deep-learning models. Additionally, we also present a comprehensive algorithm for the implementation of parallelizable SNNs and LSMs that are GPU-compatible. We implement the PLSM model to classify unintentional/accidental video clips, using the \textit{Oops} dataset. From the experimental results on detecting unintentional action in video, it can be observed that our proposed model outperforms a self-supervised model and a fully supervised traditional deep learning model. All the implemented codes can be found at our repository \url{https://github.com/anonymoussentience2020/Parallelized_LSM_for_Unintentional_Action_Recognition}.
\end{abstract}

}

\maketitle

\IEEEdisplaynontitleabstractindextext

%
\IEEEpeerreviewmaketitle

\IEEEraisesectionheading{
\section{Introduction}
\label{sec:introduction}}

To enable the deployment of cutting edge AI algorithms on low-end embedded platforms to solve real-time problems, a recent surge of interests have been channelised towards the development of bio-inspired neuromorphic systems \cite{spinnaker, deep_lsm}. These systems mimic the neuronal information processing techniques used in the human brain and allow researchers to realise neural systems directly in hardware \cite{vlsi_snn}. This offers great potential to develop computationally light algorithms for real-time problems. The neural networks used in these systems are also called Spiking Neural Networks (SNNs). Unlike the non-spiking neurons used in conventional deep networks like Convolutional Neural Networks (CNNs) and Recurrent Neural Networks (RNNs), SNNs employ neurons that communicate using spikes (mathematically represented as a Dirac delta function), instead of propagating analog values. However, the neuromorphic neuronal systems very often suffers from under-performance when compared to its deep learning counterparts. This happens especially because the activation functions (unit step functions) used in SNNs are non-differentiable, which obstructs the use of traditional backpropagation and gradient descent learning methods. This becomes even more challenging for complex tasks like activity and action recognition from video, which are performed seamlessly by fully supervised traditional deep frameworks. However, conventional deep learning frameworks (specially RNNs) are computationally expensive to train and sometimes unstable (vanishing and exploding gradient problems), which hinder their deployment on low-end embedded platforms. 

Complex task like spatio-temporal data analysis is a growing field of interest for deep learning researchers, though it has been explored less compared to image, audio and text analysis. This is mainly because of the complexity of information processing required to perform substantially on video data. Tasks like action \cite{action_recog} and activity recognition \cite{activity_recog} from video demand both spatial and temporal feature extraction, along with the abstraction of high level information. Further, inferring information like motives or intentions portrayed in a video adds another level of complexity and computational resource overhead to the algorithms. This not only hinders the training of the deep learning models, but also the deployment of the same on user-end devices \cite{deep_lsm}.

A paradigm of computational models, known as Reservoir Computing (RC) \cite{RC}, has also gained substantial attention in the recent times. The RC algorithms were initially proposed by two independent research groups as Liquid State Machine (LSM) \cite{maass} and Echo State Network (ESN) \cite{esn}. The algorithms try to bypass the problem of training highly recurrent neural networks by assuming that even a randomly connected fixed weight RNN possesses enough computational ability to project the input data to a very high dimensional sparse latent space so that data from distinct classes become almost linearly separable. The LSM model makes use of SNNs whereas the ESN model uses non-spiking neurons.

The major problem that hinders the growth of reservoir computing algorithms like LSMs, is the inability to simulate spiking neurons in a parallelizable or GPU-compatible fashion. Therefore, taking into consideration all the stated problems, we present an algorithm that parallelizes spiking neurons in general. We find that the parallelized spiking neurons function identical to the traditional spiking neurons. The algorithm is further developed to implement a parallelized LSM. Eventually, we assess the computational capability of bio-inspired neuromorphic SNNs, based on the parallelized LSM model, to predict unintentional actions in videos. In contrast to object recognition tasks in videos \cite{object_recognition}, predicting intention incorporates a harder challenge of understanding complex dynamics of video frame contents and temporal correlations among them. We use the recently released \textit{Oops} dataset \cite{oops}, which contains unconstrained in-the-wild fail/accidental videos. We devise a novel LSM architecture, the Parallelized Liquid State Machine (PLSM), for this specific task. Experiments show that our proposed PLSM model outperforms the state-of-the-art self-supervised and fully supervised model. However, we recognize the parallelization of spiking neurons/LSMs as the primary motive and contribution of this work. 

The major contributions of this work are :
\begin{enumerate}
    \item Implementation of parallelizable spiking neural networks that are GPU-compatible. To the best of our knowledge, such an implementation is done for the first time in literature.
    \item Construction of a novel LSM architecture, the PLSM model (using the developed parallelized spiking neurons formulation), that performs better than state-of-the-art self-supervised and fully supervised deep learning model at predicting unintentional action in video.
    \item Use of a novel convolutional read-out layer for the LSM that incorporates both spatial and temporal information.
    \item Design of a novel masking technique to incorporate temporal/semantic structure in the predicted output of the LSM model.
\end{enumerate}

The remainder of this paper is organized as follows: Section \ref{related_work} presents an insight into the existing literature of video analysis, the \textit{Oops} dataset and reservoir computing. In Section \ref{proposed_methodology}, we discuss the proposed methodology in detail. The dataset description and implementation details are described under Sections \ref{dataset} and \ref{problem_formulation}. In Section \ref{results}, we present the quantitative results obtained by our proposed model, followed by an ablation study in Section \ref{ablation}. Finally, Section \ref{conclusion} concludes the paper.

\begin{figure*}[ht]
\centering
     \includegraphics[width=16 cm, height=7cm]{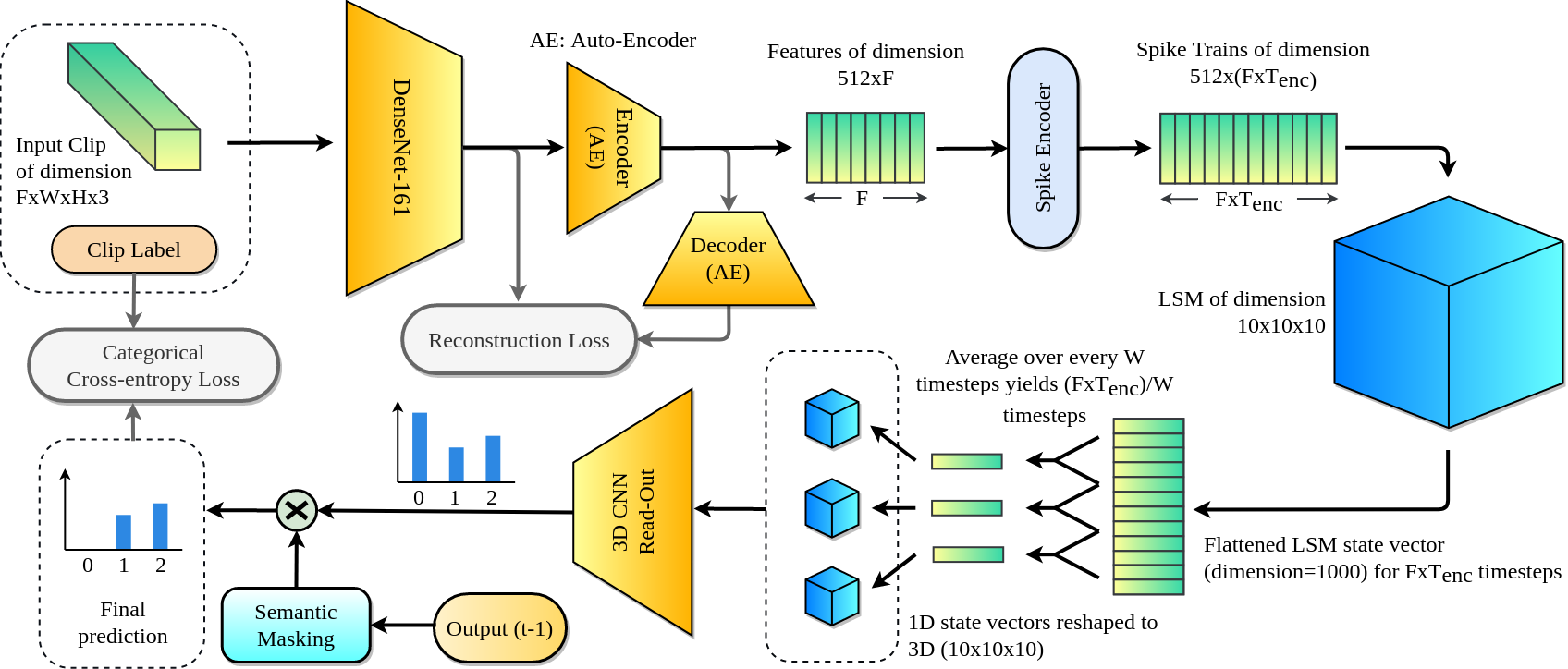} 
\caption{\textbf{The PLSM architecture:} Schematic diagram of the complete PLSM model. The LSM cube visualized in the diagram includes the input layer of the LSM as well.}
\label{fig_full_model}
\end{figure*}

\section{Related Work}

\label{related_work}

\noindent\textbf{Video analysis and action recognition:} Our primary goal of predicting accidental/unintentional actions in video is a subset of a larger domain of video analysis. Object recognition from videos \cite{object_recognition} had only been the beginning in this field. Intrinsically, objects have a nature of lucid persistence in form across video frames. However, recognizing action, that incorporates dynamic changes in form of content in a video, is a more challenging task. In earlier times, classifying action from video \cite{v1, v2, v3, v4, v5} made use of feature descriptors for given video frames. However, with the advent of deep CNNs, trainable features became the trend. Consequently, informative feature representation techniques like visual information fusion \cite{visual_fusion_1, visual_fusion_2}, 3D CNNs \cite{ST_3dcnn} and bilateral CNNs \cite{bilateral_cnn} became the state-of-the-art. Further, researchers have also proposed LSTM on visual features extracted by CNN \cite{lstm_on_cnn} and combining motion sensors with visual features \cite{motion_sensor}.

\noindent\textbf{The \textit{Oops} Dataset:} Recognizing accidental/unintentional actions in videos is a very recent sub-category of video analysis. With the release of the \textit{Oops} dataset \cite{oops}, a benchmark had been created to explore this direction. Further, this dataset contains in-the-wild videos, which are captured by amateur videographers in the real world. Exploring methodologies that would work in these scenarios would greatly help to develop models that could be deployed in real-life and real-time situations. In the baseline work \cite{oops}, the authors explore the potential of self-supervised feature extractions in detail, namely video speed, video context and event order. Using these features, three different tasks were addressed: classification, localization and anticipation. They achieve an accuracy of 61.6\%, 65.3\% and 56.7\% in the respective tasks.

\noindent\textbf{Reservoir Computing:} Recent advancements in deep learning, which are befitting for high-end computational platforms, often suffer from deployment problems when considering low-end embedded platforms. Reservoir computing (RC) \cite{RC} addresses this issue by bypassing the need for training highly recurrent neural networks or reservoirs. Liquid State Machine (LSM) \cite{maass} (uses spiking neurons) and Echo State Networks (ESN) \cite{esn} (uses non-spiking neurons) are the two primary algorithms of RC. Recently, Convolutional Drift Networks (CDN) \cite{cdn} were proposed to perform power and computational resource efficient video activity recognition. CDNs used CNNs to extract features from video frames and an ESN was used as the reservoir that captured the temporal dynamics. Additionally, \cite{deep_lsm} presented a hierarchical deep-LSM model along with attention models to solve the problem of activity recognition from egocentric video. Apart from videos, several works have used ESN and LSM for image processing \cite{image_processing}, speech recognition \cite{speech_recognition}, and even for reinforcement learning \cite{lsm_rl}. However, LSMs offer a greater opportunity than ESNs, because of the recent developments of neuromorphic hardware \cite{spinnaker}. These systems implement spiking neural networks directly in hardware, minimizing the simulation cost immensely. 

 In this work, we leverage the inherent potential of LSMs, modify the general architecture to our advantage and present a viable solution to spatio-temporal information processing. 

\section{Proposed Methodology}
\label{proposed_methodology}
To elucidate the scheme proposed in this work, we present the details of the proposed scheme in hierarchically. Firstly, we describe the neuronal dynamics of the spiking neurons, and the GPU-compatible parallelized version of the equations in Section \ref{snn}. Secondly, we describe the architecture and dynamics of the PLSM model proposed in this work in Section \ref{lsm}. The complete architecture of the proposed PLSM model is visualized in Fig. \ref{fig_full_model}.

\subsection{Spiking Neurons}
\label{snn}
We use the Leaky-Integrate-and-Fire (LIF) \cite{lif} model to implement the dynamics of spiking neurons in our work, which is described by the following equation:

\begin{equation}
\label{eq_1}
\mathit{ \tau_{m}\frac{\mathrm{d} V(t)}{\mathrm{d} t} = -(V(t)-V_{rest}) + R_{m}I(t)}
\end{equation}

\noindent where $V(t)$ is the membrane potential of a spiking neuron at time $t$, and $V_{rest}$ is the resting membrane potential that $V(t)$ decays to, at a decay rate governed by the time constant $\tau_{m}$. The membrane resistance is defined as $R_{m}$ and $I(t)$ is the instantaneous input current. When the neuron does not receive an input current ($I(t)=0$), the membrane potential decays at a rate proportional to its instantaneous potential, until it settles at $V_{rest}$. However, when $V(t)$ reaches a threshold potential $V_{th}$ (which is always greater than $V_{rest}$) on receiving successive spikes in a short time window, the neuron spikes by emitting an output voltage of magnitude $V_{spike}$, as expressed in Eq. \ref{eq_2}:
\begin{equation}
\label{eq_2}
\mathit{V(t) = \begin{cases} \mbox{$V_{spike}$,} & \mbox{if } V(t)\geq V_{th} \\ \mbox{$V(t)$,} & \mbox{otherwise} \end{cases}}
\end{equation}

\noindent The activation function (Eq. \ref{eq_2}) is conditional and non-differentiable in nature, due to which the traditional backpropagation based learning algorithm (which involves chained differentiation) cannot be deployed in spiking neural networks. Additionally, once a neuron spikes, it enters a refractory period for $\tau_{ref}$ timesteps, within which it's membrane potential stays constant at $V_{rest}$. Generally, for the purpose of numerical simulations, Eqs. \ref{eq_1} and \ref{eq_2} are implemented using classical conditional statements, as described in algorithm \ref{algo_1}. 

\begin{algorithm}[t]
\label{algo_1}
\caption{Generic implementation of a LIF neuron's state update routine}
\textbf{Input:} Instantaneous current $I(t)$ \\
\textbf{Output:} Membrane potential at time = $t$ \\

\If{refraction is False}
{
    $V(t) = V(t-1) + \left [ \frac{-V(t-1)+V_{rest}+R_{m}I(t)}{\tau_{m}} \right ] * \Delta t$ \\
    \If{$V(t) \geq V_{th}$}
    {
        $V(t) = V_{spike}$ \\
        $refraction$ = $True$ \\
        $counter$ = $\tau_{ref}$ \\
    }
}
\Else
{
    $V(t)$ = $V_{rest}$ \\
    $counter$ = $counter$ - 1 \\
    \If{counter is 0}
    {
        $refraction$ = $False$
    }    
}

\end{algorithm}

Note that, the membrane potential update routine, presented in Algorithm \ref{algo_1}, is presented for a single timestep ($\Delta$t) and for a single spiking neuron.
However, for implementing large networks that include thousands of spiking neurons, such a generic implementation makes the process of network inference highly iterative and time consuming. To tackle this problem, we reformulate the neuron state update algorithm (Algorithm \ref{algo_1}) in terms of vector/matrix operations and data type conversions. In this approach, for a layer containing L spiking LIF neurons, we compute a L-dimensional membrane voltage vector $\mathbf{V}(t)$ for the current timestep as follows:
\begin{equation}
\label{eq_3}
\mathit{\Delta \textbf{V}(t) = \frac{-\textbf{V}(t)\oplus \mathbf{V_{rest}} \oplus (\mathbf{I}(t)\otimes R_{m})}{\tau_{m}}}
\end{equation}
\begin{equation}
\mathit{\textbf{V}(t) = \textbf{V}(t) \oplus (\Delta \textbf{V}(t)\otimes \Delta t)    }
\end{equation}
\noindent where $\oplus$ and $\otimes$ represent element-wise addition and multiplication of a scaler with each element of a vector, respectively. $\mathbf{V_{rest}}$ is an L-dimensional vector where all elements are $V_{rest}$, and $\mathbf{I}(t)$ is the input current vector. To carry the information about refracting neurons across timesteps, we use a refraction counter vector $\mathbf{R_{c}}(t)$. Further, we compute a binary vector $\mathbf{R_f}$, which represents whether a neuron is in refractive state, as follows:
\begin{equation}
\mathit{\mathbf{R_f} = int(bool(\mathbf{R_{c}}(t))}    
\end{equation}
\noindent where,
\begin{equation}
\mathit{bool(x) = \begin{cases} \mbox{$True$,} & \mbox{if } x \neq 0 \\ \mbox{$False$,} & \mbox{if } x = 0 \end{cases}}\end{equation}
\begin{equation}
\mathit{int(x) = \begin{cases} \mbox{$0$,} & \mbox{if } x =$ $ $False$ \\ \mbox{$1$,} & \mbox{if } x =$ $ $True$ \end{cases}}.\end{equation}
\noindent The $bool(\cdot)$ function plays an instrumental role here in detecting zero-crossings. Depending on the states computed and stored in $\mathbf{R_f}$, we update the membrane voltages using the following equation.
\begin{equation}
\mathit{\overline{\mathbf{{V}}}(t) = [(1 - \mathbf{R_f})\odot \textbf{V}(t)] \oplus (\mathbf{R_{f}}\odot \mathbf{V_{rest}})},
\end{equation}
\noindent where, $\odot$ represents element-wise multiplication. The event $\mathbf{S}$, wherein a neuron's membrane potential ($\mathbf{V}(t)$) crosses the threshold magnitude $V_{th}$, is detected using Eq. \ref{eq_6}, where $\mathbf{[0]_{L}}$ represents a L-dimensional zero vector.
\begin{equation}
\label{eq_6}
\mathit{\mathbf{S} = int(bool([max(\mathbf{[0]_{L}}, [\overline{\mathbf{{V}}}(t) \ominus \mathbf{V_{th}}])]))},
\end{equation}
\noindent where $\ominus$ and $\mathbf{V_{th}}$ represent element-wise subtraction and a L-dimensional vector where all elements are $V_{th}$. A vector $\mathbf{N}(t)$ is computed containing the instantaneous output of the entire layer using Eq. \ref{eq_10}. Consequently, any element in $\mathbf{N}(t)$ will either be 0 or a non-zero scalar of magnitude $V_{spike}$. Therefore, a L-dimensional vector $\mathbf{V_{spike}}$, whose all elements are $V_{spike}$, is used in the computation of $\mathbf{N}(t)$.
\begin{equation}
\label{eq_10}
\mathit{\mathbf{N(t)} = \mathbf{S} \odot \mathbf{V_{spike}}}.    
\end{equation}
\noindent After each timestep, the refraction counter vector $\mathbf{R_c}(t)$ is decremented for the refracting neurons using Eq. \ref{eq_11}.
\begin{equation}
\label{eq_11}
\mathit{\mathbf{\overline{R_{c}}}(t+1) = \mathbf{R_c}(t) \ominus \mathbf{R_f}}.
\end{equation}
\noindent Finally, at the end of each timestep, $\mathbf{V}(t+1)$ and $\mathbf{R_c}(t+1)$ is calculated for the next timestep using Eqs. \ref{eq_12} and \ref{eq_13}. Neurons that have spiked in the current timestep update $\mathbf{R_c}(t)$ with a magnitude of $\tau_{ref}$, representing a transition from normal to refractory mode. After $\tau_{ref}$ timesteps, these neurons exit the refractory period.
\begin{equation}
\label{eq_12}
\mathit{\textbf{V}(t+1) = ((1-\mathbf{S}) \odot \overline{\mathbf{{V}}}(t)) \oplus \mathbf{N(t)}}.
\end{equation}
\begin{equation}
\label{eq_13}
\mathit{\mathbf{R_c}(t+1) = (\mathbf{S} \otimes \tau_{ref}) \oplus \mathbf{\overline{R_{c}}}(t+1)}.
\end{equation}

\begin{figure}[t]
\centering
     \includegraphics[width=8.3 cm]{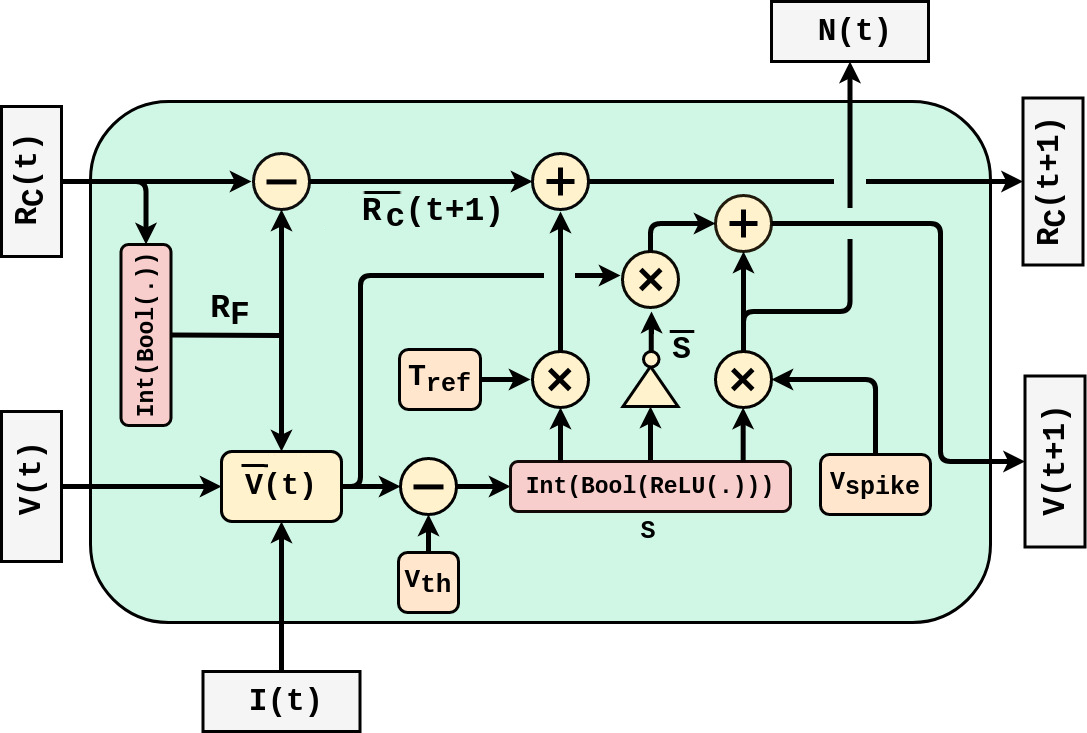} 
\caption{\textbf{Parallelized LIF layer}: Using the Eqs. \ref{eq_3}-\ref{eq_13}, a parallelized LIF layer containing L neurons can be represented in a modular fashion. }
\label{fig_lif_layer}
\end{figure}

Note that, in the above formulations, all the vectors are L-dimensional, considering a single LIF layer processing a single spike-train of dimension L$\times $T$_{ST}$, where T$_{ST}$ is the temporal length of the spike train. However, it can be easily scaled up to batch-wise prediction mode by considering L$\times$B dimensional matrices instead, for all the state vectors, where B is the input batch size (considering the parallel processing of B spike-trains). Fig. \ref{fig_lif_layer} presents a modular representation of Eqs. \ref{eq_3}-\ref{eq_13} for a LIF layer containing L neurons. 

\subsection{Liquid State Machine (LSM)}
\label{lsm}
Initially proposed by \cite{maass}, an LSM is composed of three basic parts: i) Input Layer, ii) Liquid Layer, and iii) Read-out Layer. The input layer is composed of a set of neurons that receives spike trains over time. The received spikes are then propagated to the liquid layer through a sparsely connected fixed weight matrix. The weight matrix is generated during the initialization phase of the LSM. Inside the liquid layer, the neurons are primarily divided as: primary neurons and auxiliary neurons, based upon whether a neuron receives input from input layer \cite{deep_lsm}. They are further classified as excitatory (E) and inihibitory (I) neurons, based upon whether their outputs influence other neurons positively or negatively. Only the primary excitatory neurons receive direct input from the input layer of the LSM. The input layer to liquid layer weight matrix ($W_{LI}$) is initialized in a way such that each primary excitatory neuron receives input from a sparse number of random input layer neurons. This scheme allows a uniform excitation of the liquid layer's excitatory neurons by the input spike train.

The liquid layer plays the most important role in an LSM. It is composed of a set of neurons (LIF neurons in this work), which are sparsely connected to each other following some initialization scheme. We maintain a ratio of 4:1 excitatory to inhibitory neuron ratio in the liquid layer, following \cite{lsm_rl, deep_lsm}. A liquid layer weight matrix ($W_{L}$) determines the interconnections among the liquid layer neurons. $W_{L}$ forms the recurrence within the liquid layer, and it is kept fixed during the entire experiment. In the initialization phase of the LSM, the probability that any two neurons $n_i$ and $n_j$ will have a connection is given by the following equation:
\begin{equation}
\label{lambda_rule}
    \mathit P(w_{i,j}\neq0) = Ce^{(\frac{-D(i,j)}{\lambda})^2}
\end{equation}
\noindent where C determines the maximum probability of a connection (based upon the type of neurons (E or I) and the direction of connection), $\lambda$ determines how sharply the probability decreases with distance, and $D(\cdot)$ computes the Euclidean distance between the positions of neurons $n_i$ and $n_j$. Finally, in the event of a successful connection, the absolute weight of the connection is decided based upon the type of neurons and direction of connection. The values of C, $\lambda$ and connection weights used in this work are mentioned in the supplementary material, along with other LSM and LIF neuron parameters. Further, we also use a parameter $W_{scale}$ to scale the overall weight distribution of the input and liquid weight matrix, since the distribution of synaptic strengths plays an important role in determining the dynamics of the LSM \cite{W_lambda}.

The read-out layer presents the final output of a LSM. A fully connected weight matrix ($W_{RL}$) maps the activation of all excitatory neurons in the liquid layer to the read-out layer neurons. However, in a generic LSM model, the activation of the liquid layer is compressed temporally by computing the mean spike count for each neuron. This yields a 1D vector, which is then transformed by $W_{RL}$. Depending on the type of the problem at hand, one can choose the activation function for the read-out layer neurons. Finally, to train a LSM, traditional backpropagation can be used to optimize the weights of $W_{RL}$.

The details of a LSM discussed so far considers a generic model. However, in this work we incorporate some modifications which aid our specific purpose of detecting unintentional actions in video. These are discussed in the following sub-sections. 

\begin{figure*}[ht]
\centering
    \includegraphics[width=16 cm]{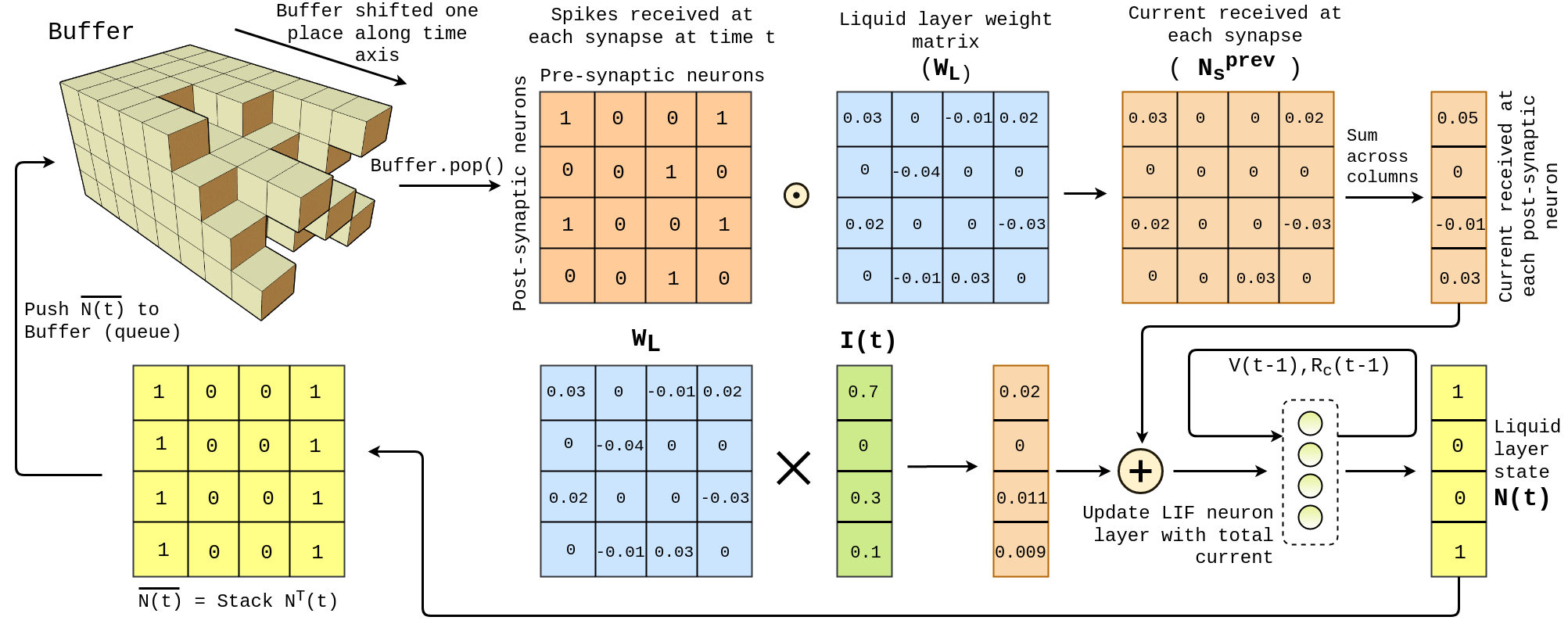} 
\caption{\textbf{LSM state update:} The LSM state update routine mentioned in algorithm \ref{algo_2} is visualized in this schematic diagram. \iffalse The $Buffer$ is shown only to the extent of the specific queue lengths for specific neuron pairs.\fi I(t) denotes the input current and N(t) represents the liquid layer's state for each timestep.}
\label{fig_lsm_routine}
\end{figure*}

\subsubsection{Synaptic delay}
In the liquid layer of a LSM, the computation at each iteration can be expressed as,
\begin{equation}
    \mathit I_{L}(t) = [W_{L} * N(t-1)] + [W_{LI} * I_{I}(t)],
\end{equation}

\noindent where, $I_{L}(t)$ is the final input current at time $t$ for all the neurons in the liquid layer, $N(t-1)$ is the previous state of the liquid layer and $I_{I}(t)$ is the instantaneous input current that arrived from the input layer. $I_{L}(t)$ is then used to update the state of the liquid layer neurons using the Eq. \ref{eq_3}-\ref{eq_13}. Therefore, in this generic implementation, it is considered that when a spike originates in any liquid layer neuron (say $n_i$), it reaches all other post-synaptic neurons (neurons that $n_i$ connects to) exactly in the next iteration. However, considering a physical environment wherein the liquid layer neurons reside, it is trivial that the time required by a signal to travel different lengths of connections will be different (since $n_i$ is at different distances from different post-synaptic neurons). This brings in the concept of connections with delay \cite{delay}. Algorithm \ref{algo_2} describes the state update routine for a liquid layer containing L spiking neurons that incorporates synaptic delay, and Fig. \ref{fig_lsm_routine} visualizes the algorithm schematically.

\begin{algorithm}[t]
\label{algo_2}
 \caption{LSM update routine with delayed synaptic connections}
Input: Input spike-train of length T \\
Output: Activation of liquid layer\\

\For{ $t \rightarrow$ \text{T}} 
   { 
   $N_{s}^{prev}$ = $W_L \odot Buffer.pop()$ \\
   $N(t-1) = \sum N_{s}^{prev}$,$ $ across $ $columns \\
   $I_{L}(t) = N(t-1) + [W_{LI} * I_{I}(t)]$ \\
   $N(t) = update(I_{L}(t))$ \\ 
   \tcp{Stack N(t) L times across columns} 
   $\overline{N(t)} = [ N^{T}(t), N^{T}(t), ..., N^{T}(t) ]$\\
   $Buffer$.push($\overline{N(t)}$) \\
   $activation[t] = N(t)$
   }
\end{algorithm}

In algorithm \ref{algo_2}, the $update(\cdot)$ function represents the state update of the liquid layer LIF neurons (using Eq. \ref{eq_3}-\ref{eq_13}), and $activation[\cdot]$ contains the final output of the liquid layer.
The foundation of algorithm \ref{algo_2} is based on the use of a Queue data structure, that we call the $Buffer$. The queue is essentially 3D, wherein the data elements are 2-dimensional (L$\times$L neuronal states), and the third axis represents time. After each timestep, the elements within the queue are shifted one place along the time axis, representing the temporal flow of spikes through inter-neuron connections. A signal element $s_{i,j}$ in $\overline{N(t)}$ represents a spike transmitted from neuron $n_j$ to $n_i$. However, the timestep at which $s_{i,j}$ reaches $n_i$ is determined by the distance between $n_i$ and $n_j$ and this is reflected in the length of the queue that carries $s_{i,j}$ along the time axis specifically. Since distance between any two neurons is unique, the queue length (and therefore the delay) for each connection is also unique. This yields an uneven pop surface. Therefore, to implement a pop operation of the queue, we use a masking technique. The mask is also a 3D matrix of dimension $L$$\times$$L$$\times$$T_{max}$, same as the dimension of the queue/$Buffer$. The mask and the pop operation using mask can be defined as:
\begin{equation}
    \mathit{Mask(i,j,t) = \begin{cases} \mbox{$1$,} & \mbox{if } D(i,j) = t \\ \mbox{$0$,} & \mbox{otherwise} \end{cases}}
\end{equation}
\begin{equation}
    \mathit{pE(t) = Mask \odot Buffer(t)},
\end{equation}
\noindent where $D(\cdot)$ computes the Euclidean distance between neuron $n_i$ and $n_j$, and $pE(t)$ is the popped element/matrix at time $t$. Also, we consider that a signal takes one timestep to traverse a unit distance. The technique of using a Queue and a masking-pop operation enables us to sample spikes in a parallelized fashion from each and every synapse, wherein the signals have traversed different lengths/suffered different delays. 

\subsubsection{Spatio-temporal read-out}
\label{conv_readout}
In contrast to generic LSM models, where spatio-temporal activation of the liquid layer is compressed to obtain a 1D mean spike count vector followed by a fully-connected network for the read-out action, we devise a 3D CNN read-out layer. Firstly, we preserve the spatial information of neurons by reshaping the obtained 1D liquid layer state vectors back to its original 3D cubic form. This is because we hypothesize that the spatial arrangement of neurons carry significant information since connections in the liquid layer are initialized using a spatial probability function (Eq.\ref{lambda_rule}). Secondly, we do not compress the activations along the time axis completely. Instead, we perform mean over the liquid layer state vectors for every $W$ timesteps, such that $T/W$ is an integer, where $T$ is the total number of timesteps for which the liquid layer activation is obtained. This creates a richer representation of the liquid layer's activation over space and time.

\subsubsection{Semantic masking}

Considering the classification problem (discussed later in Section \ref{problem_formulation}) addressed in this work, the LSM model predicts whether a given clip of video is either pre-accidental (class 0), transitional (class 1) or post-accidental (class 2). To preserve temporal information over an entire video, we use the LSM in a sliding window fashion. Therefore, the LSM predicts a class label for each clip, depending upon only the present and past inputs. This generally yields an output of the following form for a given video input:
\begin{equation}
    \mathit{Output = [2, 1, 1, 0, 2, 0, ..., 1, 2]}
\end{equation}
However, we observe that the model prediction must follow a semantic constrain in which a post-accidental label (class 2) should not be predicted before a transitional label (class 1) is predicted, and same for class 1 and 0. This inherent temporal structure of the output can be described by defining a semantically correct model output as:
\begin{equation}
    \mathit{Output = [0, 0, ..., 0, 1, 1, ..., 1, 2, 2, ..., 2]}
\end{equation}
Therefore, to enforce a semantic constrain on the LSM's output, we apply a deterministic masking technique that incorporates previous outputs of the model. The mask is defined as:
\begin{equation}
    \mathit{Mask_{sem} = \begin{cases} \mbox{$[1,1,0]$,} & \mbox{if } last $ $ prediction = 0 \\ 
                                 \mbox{$[0,1,1]$,} & \mbox{if } last $ $ prediction = 1 \\
                                 \mbox{$[0,0,1]$,} & \mbox{if } last $ $ prediction = 2 \\
                                \end{cases}}
\end{equation}
Finally, to obtain the LSM model's output, we perform:
\begin{equation}
    \mathit{Output = Mask_{sem} \odot P_{CNN}}
\end{equation}
\noindent where $P_{CNN}$ is the probability distribution output of the convolutional read-out layer over the 3 classes.

\section{Experiments}
\subsection{Dataset}
\label{dataset}

The \textit{Oops} dataset \cite{oops} used in this work is a collection of videos that contains unintentional human actions. Analysis of the dataset reveals that it contains 20,723 videos, in which 6170 are labelled videos for training, 4791 are labelled videos video for validation and the remaining are unlabeled videos. The dataset contains in-the-wild videos that comprises of diverse actions, environments and intentions. Videos of length greater than 30 seconds and lesser than 3 seconds were not considered in the experiments, since they often contain out-of-context or no-context information \cite{oops}. A more vivid description of the dataset statistics is presented in \cite{oops} and the dataset is  available at \url{https://oops.cs.columbia.edu/data/#download}.

The primary motive of this work is to present the parallelizable version of spiking neurons, that enable the simulation of large spiking neuron reservoirs like liquid state machine. The proposed PLSM architecture validates our parallelized framework on the \textit{Oops} dataset, specifically. Therefore, to verify the effectiveness of PLSM across datasets, we performed an experiment on a subset of HMDB-51 datatset (containing in-the-wild videos of human actions), comprising of 5 distinct action classes. We choose to prove the effectiveness of the proposed parallelizable SNN model for benchmark action recognition tasks. Therefore, for a fair comparison of the performance, we keep the number of classes selected from HMDB-51 dataset close to the number of classes in Oops dataset. 

\subsection{Implementation Details}
\label{problem_formulation}

The problem is formulated in a way such that each video is divided into C overlapping clips ($clip \in \mathbb{R}^{F \times W \times H \times 3}$). Following the baseline work \cite{oops}, we maintain a clip length F as 16 frames and an overlap of 12 frames between any two consecutive clips. This was done to ensure that the number of predicted labels matches the number of annotated labels for any video. Each frame of every clip was resized to $112 \times 112 \times 3$, followed by Z-score normalization. Eventually, we use a DenseNet-161 CNN model, pre-trained on the ImageNet dataset, to extract features from the clips frame-wise. Alternatively, we also experimented with ResNet-50, pre-trained on ImageNet dataset, as the primary feature extractor. However, DenseNet-161 was chosen as the final backbone since it yielded better validation accuracy (refer Table \ref{arch_ablation}). To reduce the 2208-dimensional feature vector (from DenseNet-161) to a 512 dimensions, we train another auto-encoder. The yielded feature vectors are then spike encoded using Poisson's spike encoding scheme \cite{poisson}. 

\begin{table}[th]
\renewcommand{\arraystretch}{1.2}
\setlength{\tabcolsep}{5pt}
\centering
\caption{Initialization values for LSM and LIF neuron hyperparameters \\ (I: Inhibitory neurons, E: Excitatory neurons)}
\vspace{3mm}
\begin{tabular}{|c|c|}
\hline
LSM and LIF neuron parameters                       & Values                 \\ \hline \hline
LSM cube dimensions                  & 10$\times$10$\times$10 neurons       \\
C {[}EE, EI, II, IE{]}               & {[}0.6, 1, 0.2, 0.8{]} \\
$\Lambda$                              & 6                      \\
Synaptic weight {[}EE, EI, II, IE{]} & {[}3, 2, -1, -4{]}     \\
$W_{scale}$                          & 0.01                   \\
Input layer size                     & 512                    \\
E:I neuron ratio                     & 0.8                    \\
Input feature selection density      & 0.1                    \\
Primary to auxiliary ratio           & 0.5                    \\ 
Spike encoding window size ($\tau_{enc}$)          & 50                    \\
$V_{th}$         & 0.1    \\
$V_{rest}$        & 0.0    \\
$V_{spike}$       & 1.0    \\
$\tau_m$         & 5.0    \\
$R_m$          & 10.0   \\
$\tau_{ref}$       & 1.0    \\ \hline
\end{tabular}
\label{hyperparameters}

\end{table}

We explicitly use the LSM in a sequential manner over each video, in contrast to the baseline work. In \cite{oops}, the dataset has been broken down into discrete clips of 16 frames, each associated with its respective label. The models presented are then trained on these collection of discrete clips, which are presented in a random order. However, in this work, we hypothesize that training a model in a sequential fashion over video helps to retain information from preceding clips. This increases the richness of the temporal information extracted by the LSM over multiple clips. Also, such a formulation provides the LSM with enough time to uniformly excite its liquid layer neurons over time. Therefore, the state vectors of the LSM are reset only after each video, and not after every clip. Note that, for the HMDB-51 subset dataset, we do not use the semantic masking module, since it does not require sequential class labelling. 

\begin{table}[ht]
\renewcommand{\arraystretch}{1.2}
\setlength{\tabcolsep}{8.4pt}
\centering
\caption{\textbf{Classification Accuracy:} We compare the validation accuracy obtained by our PLSM model and the baselines presented in \cite{oops} for the \textit{Oops} dataset.}
\vspace{3mm}

\begin{tabular}{|c|c|c|}
\hline
                                    & Method                              & Accuracy         \\ \hline \hline
\multirow{5}{*}{Baseline \cite{oops}} & Kinetics Supervision                & 64.0 \%          \\ \cline{2-3} 
                                    & Video Speed    & 61.6 \%          \\
                                    & Video Context  & 60.3 \%          \\
                                    & Video Sorting  & 60.2 \%          \\ \cline{2-3} 
                                    & Motion Magnitude                    & 44.0 \%          \\ \hline \hline
Ours                                & PLSM                                & \textbf{66.3 \%} \\ \hline
\end{tabular}
\label{table_result}
\end{table}

Finally, we use a categorical cross-entropy loss function (Eq. \ref{eq_loss}), along with learning-rate schedular, to optimize the weights of the convolutional read-out layer and train the network for 500 epochs. The performance of the PLSM model proposed in our work is subject to careful initialization of network parameters. We provide the specific values of the LSM and LIF neuron hyperparameters used in our work. Using the specified values mentioned in Table \ref{hyperparameters} shall ensure smooth reproducibility of the work.


\begin{equation}
\label{eq_loss}
    \mathit{\mathbb{L}(y,\hat{y}) = -\sum_{i=0}^{n}y_ilog\hat{y}}.
\end{equation}

\begin{table*}[t]
\centering
\caption{\textbf{Architectural ablation:} Ablation of different modules using 7 different setups for the \textit{Oops} dataset.}

\setlength{\tabcolsep}{5pt}
\renewcommand{\arraystretch}{1.3}
\begin{tabular}{lccccccc}
\hline
\multicolumn{1}{c}{Modules}                    & \textit{1st} & \textit{2nd} & \textit{3rd} & \textit{4th} & \textit{5th} & \textit{6th} & \textit{7th} \\ \hline
Backbone : DenseNet-161                        & --      & --       & \checkmark            & --       & \checkmark            & --             & \checkmark            \\
Backbone : ResNet-50                           & \checkmark            & \checkmark            & --       & \checkmark            & --       & \checkmark            & --       \\
Auto-Encoder for dimension reduction           & --       & \checkmark            & \checkmark            & \checkmark            & \checkmark            &\checkmark            & \checkmark            \\
Using mean spike count as reservoir activation & \checkmark            & \checkmark            & \checkmark            & --       & --       &--              & --       \\
3D-CNN on spatio-temporal activation           & --       & --       & --       & \checkmark            & \checkmark            & \checkmark            & \checkmark            \\
Semantic Masking                               & --       & --       & --       & --       & --       & \checkmark            & \checkmark            \\ \hline
Validation accuracy                            & \textbf{39.3}\%       & \textbf{45.7}\%       & \textbf{49.5}\%       & \textbf{56.2}\%       & \textbf{57.6}\%       & \textbf{64.3}\%       & \textbf{66.3}\%       \\ \hline
\end{tabular}
\label{arch_ablation}

\end{table*}

\subsection{Performance Comparison}
\label{results}
The primary goal of this work is to: i) validate the parallelized formulation of spiking neurons to simulate large spiking reservoirs; ii) assess the potential of spike-based bio-inspired reservoir computing algorithm, namely LSM, to solve complex problems in real-life and real-time scenarios like detecting unintentional action in realistic video. In this perspective, we compare our results obtained using the PLSM model with the baselines presented in \cite{oops}.

Table \ref{table_result} presents our result obtained on the \textit{Oops} validation dataset. The self-supervised methods proposed in \cite{oops} use a Resnet3D model (pre-trained on Kinetics action recognition dataset) to extract features from video clips. These features are then used to train a self-supervised model to predict video speed, context and order. Later, these models are used as feature extractors to classify clips using a linear classifer. On the contrary, our PLSM model extracts features using a DenseNet-161 model (pre-trained on ImageNet database) and the liquid layer of the PLSM. Therefore, only the convolutional read-out layer is imposed to training, reducing training cost. 
Additionally, we also present a result from \cite{oops}, which is obtained using simple motion detection (Motion Magnitude). For this, histogram of optical flow is computed over the videos. A multi-layer perceptron model is then trained on these features. 
Finally, we mention the result obtained by a fully supervised model (Kinetics Supervision) pre-trained on the full annotated Kinetics action recognition dataset. 

Experiments prove that our parallelized formulation of spiking neurons is functional and the PLSM model outperforms all the mentioned methodologies in terms of validation accuracy. We hypothesize that the PLSM's superior performance is obtained mainly because: i) we use the model in a sequential fashion over an entire video, retaining temporal information from previous clips, and ii) we use semantic masking to constrain the model's output to predict temporally/semantically correct labels. This enables the use of prior knowledge about the output sequence. Moreover, the dynamics of a LSM is well suited for modelling complex spatio-temporal data \cite{deep_lsm, lsm_rl}. However, using the LSM states as extracted features need a fine-grain read-out layer which acts as an interpreter. The 3D CNN read-out layer, proposed in this work, serves this purpose and enhances the overall performance of the PLSM model. 

Also, the PLSM achieves a validation accuracy of 62.5\% on the HMDB-51 subset dataset. This proves that the parallelized spiking neurons, constituting the PLSM reservoir, function appropriately across datasets.

\subsection{Ablation Study}
\label{ablation}

The results obtained by our PLSM model on the \textit{Oops} dataset are highly subject to specific module selection and network hyper-parameter selections. Table \ref{arch_ablation} presents the ablation of different modules used in our overall architecture. Additionally, we experiment with different hyper-parameters of significant parts of the architecture to justify our proposed model. 

\begin{figure}[t]
\renewcommand{\arraystretch}{0.8}
\centering
    \includegraphics[height=5cm, width=8cm]{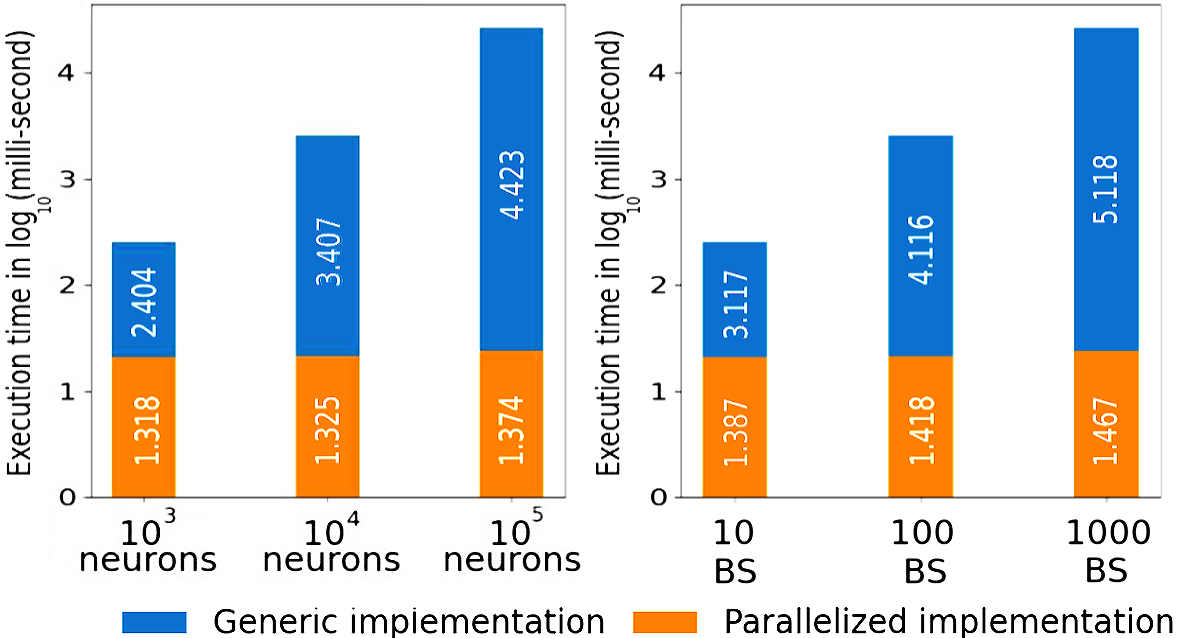}
\caption{\textbf{Comparative results of parallelization:} Our parallelized LIF neurons outperform generic LIF neurons by several orders of magnitude for a single layer with increasing neurons (left). We also implement a layer of 500 LIF neurons to evaluate the performance in batch-mode of different sizes (BS) (right).}
\label{fig_snn_layer}
\end{figure}

\textbf{Parallelization:} We evaluate the significant leverage obtained using parallelized spiking neural networks in contrast to generic implementation. Fig. \ref{fig_snn_layer} (left) outlines the significant gain in processing speed for a layer of LIF spiking neurons. As mentioned in Section \ref{snn}, our implementation of parallelized SNN layers can be easily upgraded to perform in batch-wise prediction mode. This is achieved by adding another dimension to all the state vectors of the SNN layer. The gain in processing speed while using batch-wise prediction mode for a layer of 500 LIF neurons is plotted in Fig. \ref{fig_snn_layer} (right). The results are presented for a input spike train having 100 timesteps. It is clear from the plots that with increasing number of neurons/batch-size, our parallelized SNN implementation outperforms the generic implementation algorithm with a huge margin. It is because of this computational speed that our PLSM model was able to train on a considerably big dataset like \textit{Oops} in tractable amount of time.

\begin{figure}[t]
\centering
    \includegraphics[height=4.4cm, width=8cm]{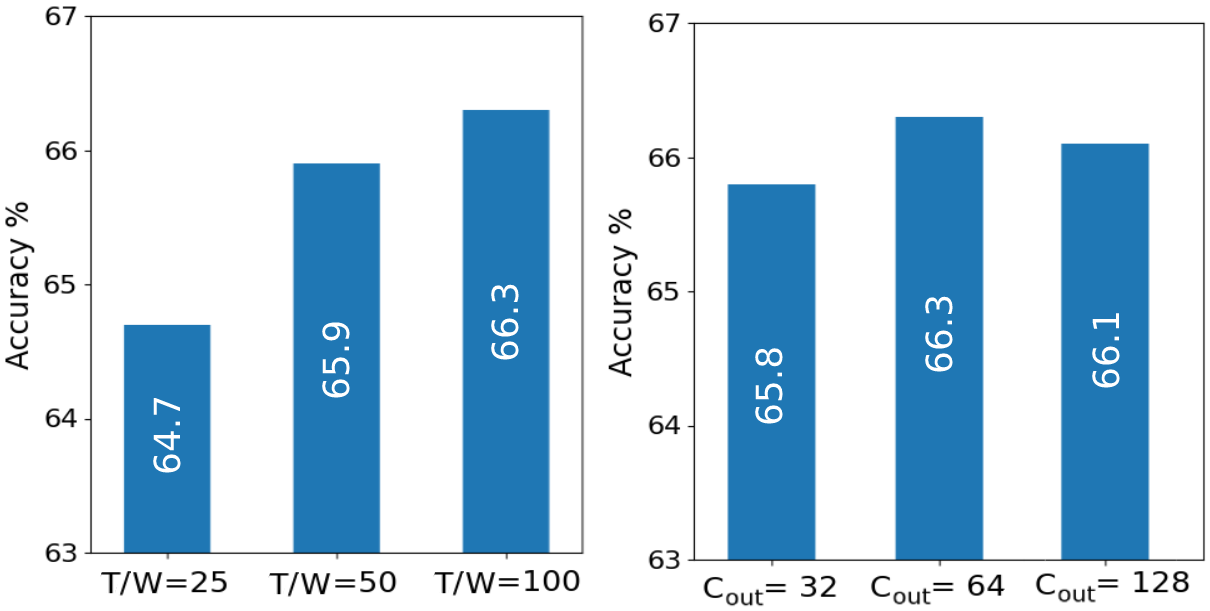} 

\caption{\textbf{Temporal precision and $\mathbf{C_{out}}$:} We find that preserving features across time increases PLSM's performance (left). For $T/W$=100, experiments show that $C_{out}$=64 provides the optimum result (right).}

\label{fig_temporal_precision}
\end{figure}

\vspace{1mm}
\textbf{Temporal precision:} We use a 3D convolutional read-out layer to classify clips from features extracted by the PLSM's liquid layer. Conventionally, the features obtained from the liquid layer are averaged over the time axis to obtain a mean spike count. However, the use of 3D CNN enables us to use the temporally distributed features by stacking them across the channel axis of the CNN's input. Fig. \ref{fig_temporal_precision} (left) shows the variation in classification accuracy with respect to different values of $T/W$ (refer Section \ref{conv_readout}). It is observed that preserving features across time enhances the overall performance.

\textbf{3D CNN output channel size:} Our convolutional read-out layer consists of a single 3D convolutional layer, followed by dropout, 3D max-pooling and a fully connected layer. We experimented with multiple cascaded 3D convolutional layers, but increasing the number of trainable parameters caused overfitting. Therefore, a single 3D convolutional layer was used, which learns $C_{out}$ number of kernels to detect features from 3D liquid layer state vectors. For an ablation, we evaluate different values for $C_{out}$ and present our result in Fig. \ref{fig_temporal_precision} (right). We find that the best result is obtained for $C_{out}$ = 64.

\textbf{Semantic Masking:} Lastly, we evaluate the semantic masking module added at the end of the 3D CNN read-out layer, which plays an instrumental role. A classification accuracy of 53.3\% is achieved without the masking. This proves that the masking improves the PLSM model's performance by almost 13\%. 

\section{Conclusion}
\label{conclusion}
This work investigates the potential of bio-inspired spike-based reservoir computing algorithm, namely LSM, as an alternative to traditional deep learning models for achieving computationally light deployment of AI algorithms on low-end embedded systems. A novel LSM architecture, the PLSM, is proposed for this purpose, that achieves significant results on a task of unintentional action detection in realistic videos. We also provide a comprehensive formulation of implementing SNNs and LSMs in a GPU-compatible parallelized fashion. Even though there remains further scope for improvement of performance, this work paves the way towards future development of neuromorphic systems.


%

\ifCLASSOPTIONcaptionsoff
  \newpage
\fi



%

{\small
\bibliographystyle{unsrt}
\bibliography{main_file}
}

%

\vskip 0pt plus -1fil

\begin{IEEEbiographynophoto}{Dipayan Das}
received his B.Tech degree in Electronics and Communication Engineering from National Institute of Technology Durgapur, India in 2020. He is currently researching as a Project Linked Person at Computer Vision and Pattern Recognition Unit, Indian Statistical Institute, Kolkata, India. His research interests include cognitive computing architectures, computer vision, robotics and artificial general intelligence. 
\end{IEEEbiographynophoto}

\vskip 0pt plus -1fil

\begin{IEEEbiographynophoto}{Saumik Bhattacharya}
is an assistant professor in the Department of Electronics and Electrical Communication Engineering, Indian Institute of Technology Kharagpur. He received the B.Tech. degree in Electronics and Communication Engineering from the West Bengal University of Technology, Kolkata in 2011, and the Ph.D. degree in Electrical Engineering from IIT Kanpur, Kanpur, India, in 2017. His research interests include image processing, computer vision, and machine learning.
\end{IEEEbiographynophoto}
\vskip 0pt plus -1fil
\begin{IEEEbiographynophoto}{Umapada Pal}
received his Ph.D. in 1997 from Indian Statistical Institute, Kolkata, India. He did his Post Doctoral research at INRIA (Institut National de Recherche en Informatique et en Automatique), France. From January 1997, he is a Faculty member of Computer Vision and Pattern Recognition Unit of the Indian Statistical Institute, Kolkata and at present he is Professor and former Head of Computer Vision and Pattern Recognition Unit. His fields of research interest include Digital Document Processing, Optical Character Recognition, Biometrics, Word spotting, Video Document Analysis etc.
\end{IEEEbiographynophoto}
\vskip 0pt plus -1fil
\begin{IEEEbiographynophoto}{Sukalpa Chanda} 
is currently working as an Associate Professor in the Department of Information Technology of Østfold University College in Norway since July 2019. Prior to this, he was appointed as a Post-Doctoral Researcher in the “Centre for Image Analysis” at the Department of Information Technology of Uppsala University, Sweden. Before he joined Uppsala University, he was engaged as a Post-Doctoral researcher for two years in the Bernoulli Institute for Mathematics, Computer Science and Artificial Intelligence, Faculty of Science and Engineering Engineering at University of Groningen, in The Netherlands. He obtained his Ph.D. in Computer Science in June 2015, from NTNU i Gjøvik, Norway. His current research interest includes  Zero-shot learning, Representation Learning  applications in  different forms of  Image Analysis problem (Document, Video). He  is as an IEEE member. 
\end{IEEEbiographynophoto}

\end{document}